\documentclass[lettersize,journal]{IEEEtran}
\usepackage{amsmath,amsfonts}
\usepackage{algorithmic}
\usepackage{algorithm}
\usepackage{array}
\usepackage[caption=false,font=normalsize,labelfont=sf,textfont=sf]{subfig}
\usepackage{textcomp}
\usepackage{stfloats}
\usepackage{url}
\usepackage{verbatim}
\usepackage{graphicx}
\usepackage{cite}
\usepackage{balance}
\usepackage{authblk}
\usepackage{amssymb}
\usepackage{booktabs}
\usepackage{bbding}
\usepackage{hyperref}
\usepackage{ifpdf}
\usepackage{xspace}
\usepackage{tabularx}
\usepackage{multirow}
\usepackage{booktabs}
\usepackage{longtable}
\usepackage{xcolor}
\usepackage{pifont}
\usepackage{textcomp}
\usepackage{tikz}
\newcommand{\circled}[1]{\tikz[baseline=(char.base)]{
    \node[shape=circle,draw,inner sep=1pt] (char) {#1};}}
\newcolumntype{Y}{>{\raggedright\arraybackslash}X} 
\newcolumntype{Z}{>{\raggedleft\arraybackslash}X}
\newcolumntype{W}{>{\centering\arraybackslash}X}

\begin{document}

\title{Neuron Abandoning Attention Flow: Visual Explanation of Dynamics inside CNN Models}

\author{Yi Liao, Yongsheng Gao,~\IEEEmembership{Senior Member,~IEEE}, Weichuan Zhang,~\IEEEmembership{Member,~IEEE}
\thanks{This work is supported in part by the Australian Research Council under Industrial Transformation Research Hub Grant IH180100002 and Discovery Grant DP180100958 (Corresponding author: Yongsheng Gao)

Yi Liao and Yongsheng Gao are with the School of Engineering and Built Environment, Griffith University, Brisbane, Queensland, 4111, Australia (e-mail: yi.liao2@griffithuni.edu.au; yongsheng.gao@griffith.edu.au). 

Weichuan Zhang is with the Institute for Integrated and Intelligent Systems, Griffith University, Brisbane, Queensland, 4111, Australia (e-mail: weichuan.zhang@griffith.edu.au) and also with the School of Electronic Information and Artificial Intelligence, Shaanxi University of Science and Technology, Xi'an 710026, Shaanxi Province, China (e-mail: zwc2003@163.com).
}
}

\markboth{IEEE TRANSACTIONS ON PATTERN ANALYSIS AND MACHINE INTELLIGENCE}%
{Shell \MakeLowercase{\textit{et al.}}: A Sample Article Using IEEEtran.cls for IEEE Journals}


\maketitle

\begin{abstract}
In this paper, we present a Neuron Abandoning Attention Flow (NAFlow) method to address the open problem of visually explaining the attention evolution dynamics inside CNNs when making their classification decisions. A novel cascading neuron abandoning back-propagation algorithm
is designed to trace neurons in all layers of a CNN that involve in making its prediction to address the problem of significant interference from abandoned neurons. Firstly, a Neuron Abandoning Back-Propagation (NA-BP) module is proposed to generate Back-Propagated Feature Maps (BPFM) by using the inverse function of the intermediate layers of CNN models, on which the neurons not used for decision-making are abandoned. Meanwhile, the cascading NA-BP modules calculate the tensors of importance coefficients which are linearly combined with the tensors of BPFMs to form the NAFlow. Secondly, to be able to visualize attention flow for similarity metric-based CNN models, a new channel contribution weights module is proposed to calculate the importance coefficients via Jacobian Matrix. The effectiveness of the proposed NAFlow is validated on nine widely-used CNN models for various tasks of general image classification, contrastive learning classification, few-shot image classification, and image  retrieval.
\end{abstract}

\section{Introduction}
\IEEEPARstart{C}{onvolutional} neural networks (CNN) have been widely applied in various image recognition tasks~\cite{refsegmentation1,refsegmentation2,refMatchingNet,refPNet}. The interpretability of CNN models involves explaining what regions on images are looked at by the models in making their classification decisions. Visualizing the regions on images used for decision-making will make the model's prediction more trustworthy and transparent. To this end, various visual explanation methods~\cite{refDeepLIFT,refIntegratedGradient,refRISE,refABN,refCAM,refGradCAM,refLayerCAM,refScoreCAM,refRelevanceCAM} are developed to display what regions on images are used by a CNN in making its final classification decision. 
\begin{figure}[!t]
\centering
\includegraphics[width=1.0\columnwidth]{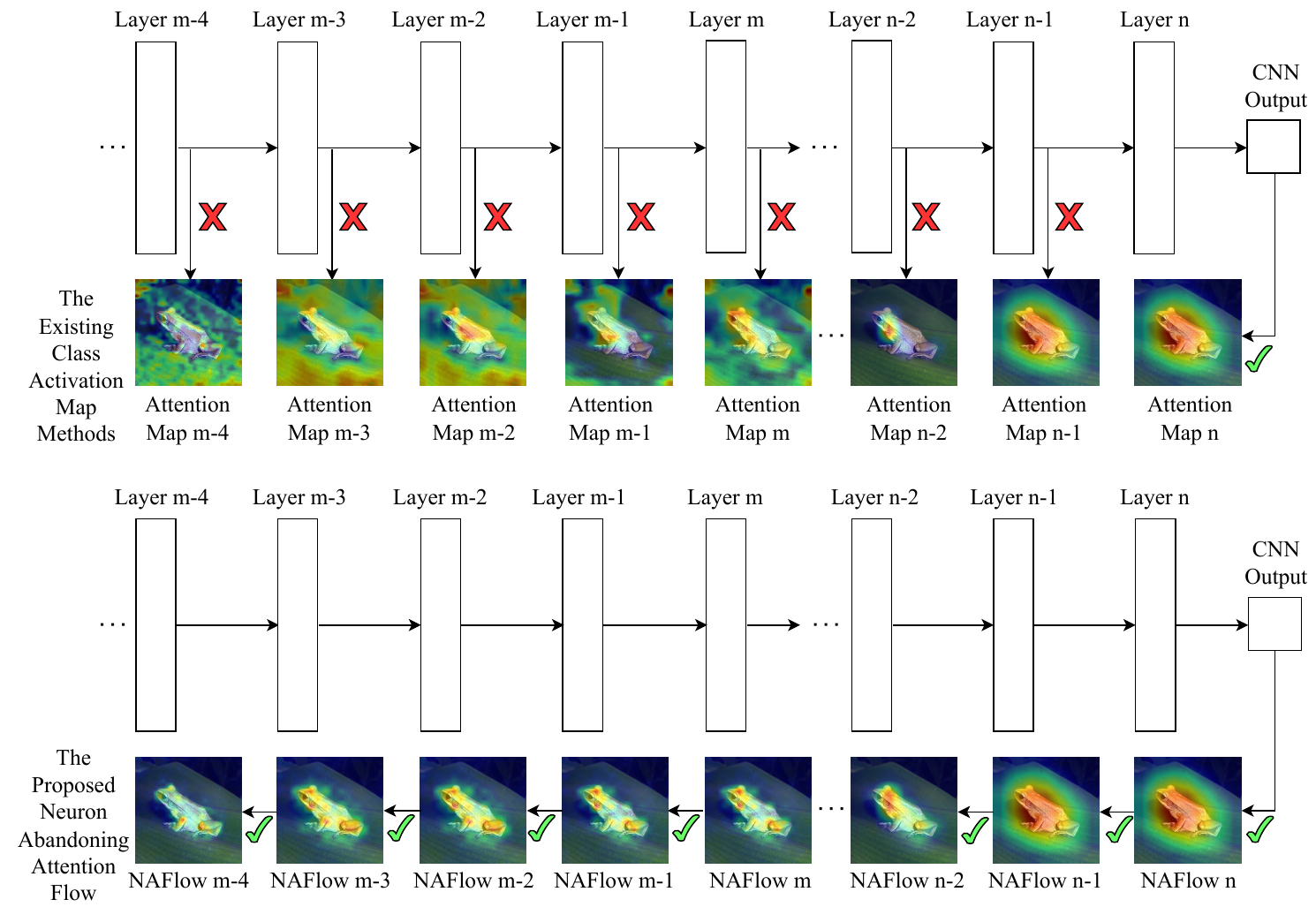}
\caption{The attention maps generated by applying the existing class activation map methods to intermediate layers in a CNN model is invalid (\textcolor{red}{$\times$} mark). The attention maps in the proposed NAFlow are generated in a neuron abandoning back propagation manner via the inverse operations of the intermediate layers in a CNN model (\textcolor{green}{$\surd$} mark indicates a valid attention map for each layer).}
\label{fig_1}
\end{figure}
However, visualizing explanation maps for not only the final layer but also the earlier intermediate layers of CNN models can reveal the evolution process of attention regions, which will be important to further unlocking the mystery hidden inside the inner structure of networks, as a tool, to better understand how a CNN progressively makes its decision or improve a design of CNN. Although designed for visualizing the saliency map of output layer of a model, Grad-CAM, Layer-CAM, and Relevance-CAM have been applied to generate the saliency maps for the intermediate layers of CNN models~\cite{refLayerCAM,refRelevanceCAM}. However, the feature maps from the intermediate layers are extracted during forward propagation (from the earlier shallow layers to the final layer of the CNN model), as shown in Fig.~\ref{fig_1} (top row). In this way, many neurons~\cite{PolarStream} on the intermediate feature maps have valid values but are not used in the final classification decision (we call these neurons the abandoned neurons) because they are, for example, either deselected by the max pooling layers, not used in convolution operations or reset to zero by ReLU activation function layers without going to the next layer (see examples in Fig.~\ref{figBPFMCNN}). They should be particularly identified and excluded from generating the explanation maps for intermediate layers. None of the existing visual explanation methods have a mechanism to remove these abandoned neurons. Thus they can only be used for generating the attention map of the final output layer but not for the intermediate layers of CNN models. Using them to generate attention maps for the intermediate layers (see Fig.~\ref{fig_1} (top row)) is conceptually incorrect with significant interference from irrelevant neurons not contributing to the decision-making of the CNN.

\begin{figure*}[t]
\centering
\includegraphics[width=1.0\textwidth]{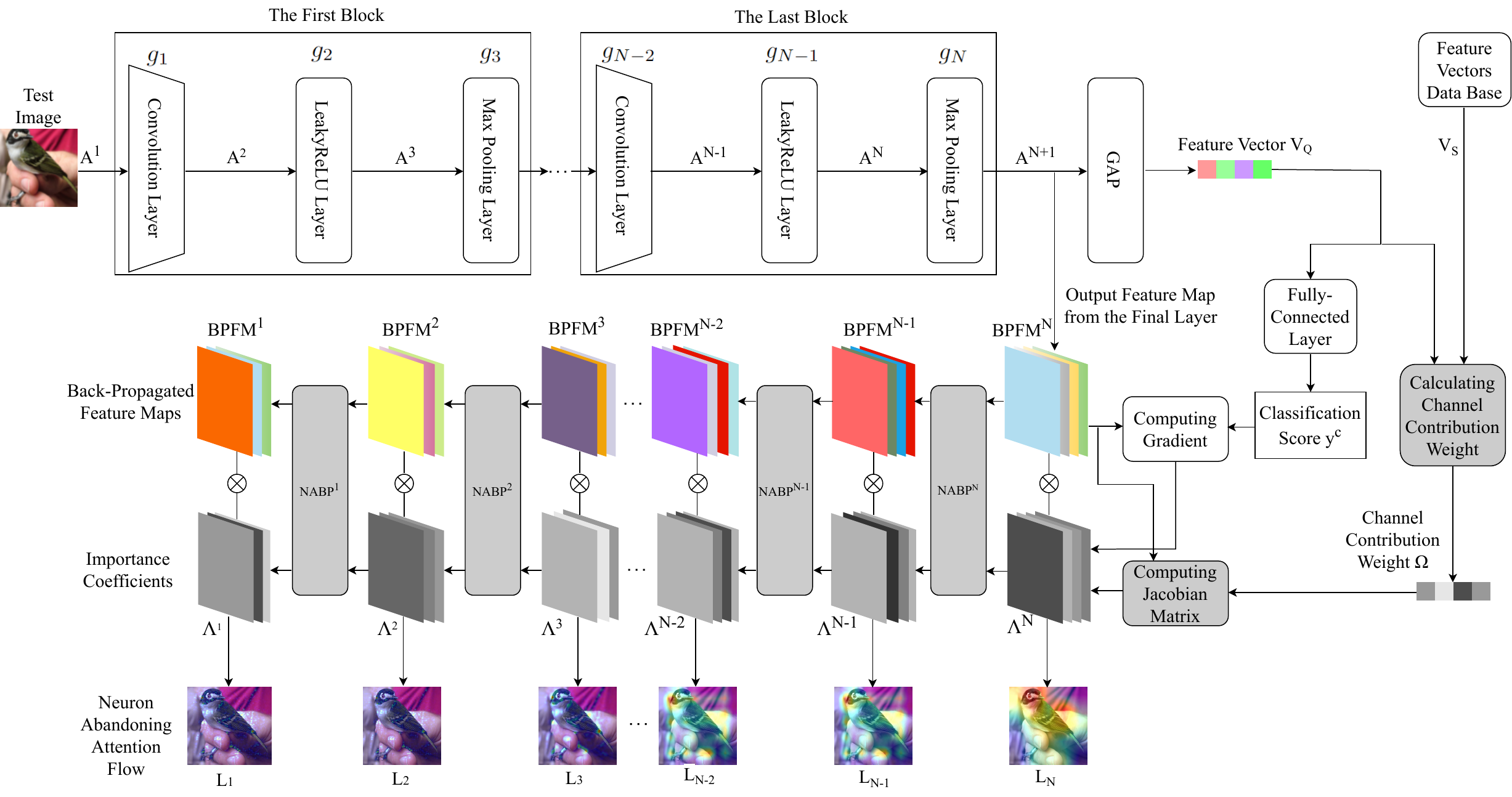} 
\caption{The pipeline of the proposed Neuron Abandoning Attention Flow in explaining a CNN model's decision. The $l$-th Back-Propagated Feature Map and the $l$-th tensor of importance coefficients are generated by the cascading Neuron Abandoning Back-Propagation (NA-BP), where $l=1,...,N$.}
\label{fig1}
\end{figure*}

In this work, we propose a backward propagation solution, for the first time, to be able to correctly visualize the inner attention map evolution/change hidden inside the CNN models. A novel Neuron Abandoning Attention Flow (NAFlow) is proposed to generate the attention maps of all layers in a CNN model. Firstly, the output feature map from the final output layer is backward propagated using the inverse functions of the intermediate layers to generate Back-Propagated Feature Maps (BPFMs), as shown in Fig.~\ref{fig1}. During this process, all the abandoned neurons are removed. Secondly, a sequence of tensors of importance coefficients are calculated by cascading Jacobian Matrix operation.  These two functions are archived by the proposed Neuron Abandoning Back-Propagation (NA-BP). Finally, the NAFlow is obtained by linearly combining the tensors of importance coefficients and BPFMs. The main contribution of this work can be summarized as follows:
\begin{itemize}
\item{We proposed a novel NAFlow method that can, for the first time, correctly visualize the attention map flow (i.e., the evolution of attention maps) inside CNN models moving from the shallow layers to the deep layers until the final output layer during the process of their decision-making.}
\item{We proposed a new back-propagation strategy to trace neurons in all layers of a CNN that involve in making its prediction. A novel cascading neuron abandoning back-propagation algorithm is designed to address the problem of significant interference from irrelevant neurons not contributing to the decision-making.}
\item{To address the incapability of existing methods in explaining CNNs using similarity metric based classification (such as CNNs for few-shot image classification, contrastive learning classification, and image retrieval), a channel contribution weight method is proposed that can visualize not only the attention map of such CNNs' output layer for the first time, but also the attention maps for all internal layers.}
\item{Extensive experimental results on nine CNNs for various tasks including general image classification, contrastive learning classification, few-shot image classification, and image retrieval demonstrate the effectiveness of the proposed method.}
\end{itemize}
\section{Background}
The existing visual explanation methods for interpreting CNN models can be broadly categorized into class-agnostic methods~\cite{refPerturbation,refRISE,refFullGradient} and class specific methods~\cite{refCAM,refGradCAM, refLayerCAM,refGradCAMplusplus,refXCAM,refRelevanceCAM,refScoreCAM, refLIFTCAM}. The class-agnostic methods generate similar visualization results regardless of the class of the image that we want to visualise~\cite{refTrasformerAttr}. They include gradient-based methods~\cite{refFullGradient} and perturbation based methods~\cite{refPerturbation,refRISE}. FullGrad~\cite{refFullGradient} generates the explanation map by assigning the importance scores to both the input features and and individual feature detectors in the networks. RISE~\cite{refRISE} estimates the region importance by probing the model with randomly masked input images and obtaining the corresponding outputs. 

Compared with class-agnostic methods, class specific methods generate class-discriminative attention maps. Class activation map (CAM)-based methods are widely applied to various downstream vision tasks (e.g., image segmentation~\cite{weaklySuperSeg} and object localization~\cite{weaklySuperLocal}) because of their high efficiency and  class-sensitive visualization. CAM-based methods focus on looking for the appropriate channel-wise importance coefficients combined linearly with the feature map that the CNN model extracts during forward-propagation. The attention maps are obtained by upsampling the combination of the importance coefficients and feature map via interpolation. CAM~\cite{refCAM} uses the weights of the FC layer (which is the classifier of the model) as the importance coefficients. Gradient-based CAM methods~\cite{refGradCAM, refLayerCAM,refGradCAMplusplus,refXCAM} obtain the importance coefficients by calculating the gradient of the classification score with respect to the feature map. Relevance-CAM~\cite{refRelevanceCAM} obtains the coefficients by assigning the negative relevance score to the non-target class and the positive relevance score to the target-class. Score-CAM~\cite{refScoreCAM} derives importance of activation maps from the contribution of highlighted input features to the model output. 

All the current methods can only be used for visualizing the attention map of final output layer of a CNN model, but not for any of the intermediate layers because the feature maps utilized by these methods are extracted during the forward propagation and all neurons (including those are not involved in the decision making) are used for generating the attention map, resulting incorrect explanation if applied on internal layers. Fig. 1 (top row) shows an example of applying Grad-CAM~\cite{refGradCAM} to visualizing internal attention map evolution in ResNet18~\cite{refResNet} when making its classification decision on ImageNet2012.

\section{Proposed Method}
In this paper, we propose a novel Neuron Abandoning  Attention Flow (NAFlow) method that can for the first time visualise how the attention map evolves inside a CNN model when making its classification decision. The overview of our method is illustrated in Fig.~\ref{fig1}. Let $f^c$ denote a CNN model. For a given image $I$, the classification score $y^c$ of the target category $c$ is obtained by the following equation,
\begin{equation}
\begin{aligned}
\label{eq1}
y^c=f^c(I).
\end{aligned}
\end{equation}
The attention map $L_{l}^c$ from the $l$-th layer in CNN model $f^c$ containing $N$ layers can be calculated by,
\begin{equation}
\begin{aligned}
\label{faf}
L_{l}^c=\text{Max}\left(\sum_{d=1}^{D^l}{{(\Lambda}^l \otimes \text{BPFM}_d^l)}, 0\right),~l=1,...,N.
\end{aligned}
\end{equation}
where ${\Lambda}^l$ and $\text{BPFM}_d^l$ denote the importance coefficients and the back-propagated feature maps for the $l$-th layer respectively, $\otimes$ denotes element-wise multiplication and $D^l$ is the number of channels from the $l$-th layer. $\text{BPFM}_d^l$ 
and ${\Lambda}^l$ are calculated via the proposed Neuron Abandoning Back-Propagation algorithm, which will be detailed in the following subsections. 
\subsection{Neuron Abandoning Back-Propagation (NA-BP)}
To simplify notation, we denote the $l$-th layer of a CNN model as $g_l$, and its input feature map is denoted as $A^l \in R^{D^l \times H^l \times W^l}$, the output feature map  $A^{l+1}$ can be obtained by the following equation, 
\begin{equation}
\begin{aligned}
\label{l-layer}
A^{l+1}=g_l(A^l),
A^{l+1}\in R^{N^{l+1} \times H^{l+1} \times W^{l+1}}.
\end{aligned}
\end{equation}
 $A^l$ contains $p$ neurons and $A^{l+1}$ contains $q$ neurons. Hence, $A^l$ and $A^{l+1}$ can be flattened by
\begin{equation}
\begin{aligned}
\label{flatten}
&[x_1,x_2,\dots,x_p]=\text{Flatten}(A^l),\\
&[y_1,y_2,\dots,y_q]=\text{Flatten}(A^{l+1}).
\end{aligned}
\end{equation}
In the following, we introduce how to obtain $[x_1,\dots,x_p]$ by using $[y_1,\dots,y_q]$ for various layers in a CNN model. 
\subsubsection{Back-Propagated Feature Map for Convolutional Layer}
\label{conv}
When $g_l$ is a convolution layer, Eq.~\ref{l-layer} and Eq.~\ref{flatten} can be expressed as
$\begin{bmatrix}
y_1\\
\vdots\\
y_q
\end{bmatrix}=\begin{bmatrix}
w_{11} & w_{12} & \cdots & w_{1p} \\
\vdots & \vdots & \ddots & \vdots \\
w_{q1} & w_{q2} & \cdots & w_{qp}
\end{bmatrix}\begin{bmatrix}
x_1\\
\vdots\\
x_p
\end{bmatrix}+\begin{bmatrix}
b_1\\
\vdots\\
b_q
\end{bmatrix}$, where $[b_1,\dots,b_q]$ denotes the bias provided by the convolution layer, 
$w_{ij}$ denotes the weights between $y_i$ and $x_j$ calculated by using Jacobian Matrix $\frac{\partial(y_1,\dots,y_q)}{\partial(x_1,\dots,x_p)}$, which is
\begin{equation}
\begin{aligned}
\label{JacobiMatrix}
w_{ij}=\frac{\partial{y_i}}{\partial{x_j}},~i=1,\cdots,q,~j=1,\cdots,p.
\end{aligned}    
\end{equation}
Because the convolution operation can be partially-connected, $w_{ij}$ will be $0$ if there is no connection between $y_i$ and $x_j$. 

In our design, for a convolutional layer $g_l$ with $q \ge p$,  we remove $[y_{p+1},\dots,y_q]$ and obtain our back-propagated feature map by
\begin{equation}
\begin{aligned}
\label{BPFM}
\text{BPFM}_l&=\text{Reshape}([x_{1},\cdots,x_{p}]),\\
\begin{bmatrix}
x_{1}\\
\vdots\\
x_{p}
\end{bmatrix}=&\left(\begin{bmatrix}
w_{11} & w_{12} & \cdots & w_{1p} \\
\vdots & \vdots & \ddots & \vdots \\
w_{p1} & w_{p2} & \cdots & w_{pp}
\end{bmatrix}^{-1}\begin{bmatrix}
y_1-b_1\\
\vdots\\
y_p-b_p
\end{bmatrix}\right). 
\end{aligned}
\end{equation}
For a convolutional layer $g_l$ with $q < p$,  we consider multiple layers $g_l$, $g_{l+1}$, ..., $g_{l+M-1}$ by increasing $l$  until the number of neurons on the output of $g_{l+M-1}$ becomes bigger than or equal to $p$, for the purpose of calculating Jacobian Matrix. Then our back-propagated feature map is computed by
\begin{equation}
\begin{aligned}
\label{BPFM2}
\text{BPFM}_l&=\text{Reshape}([x_{1},\cdots,x_{p}]),\\
\begin{bmatrix}
 x_{1}\\
\vdots\\
x_{p}   
\end{bmatrix}=&\left(\begin{bmatrix}
\frac{\partial(y_{1}^{l+M-1},\dots,y_{q}^{l+M-1})}{\partial(x_{1}^{l},\dots,x_{p}^{l})}
\end{bmatrix}^{-1} \begin{bmatrix}
y_1^{l+M-1}-{y_{1}^{l+M-1}}_{|X=0}\\
\vdots\\
y_p^{l+M-1}-{y_{p}^{l+M-1}}_{|X=0}\\
\end{bmatrix}\right).
\end{aligned}    
\end{equation}

\textit{Proof}: Condition 1: When $q \ge p$, the Jacobian Matrix is a full column rank matrix, whose rank is $p$. This means that we only need $p$ linearly independent neurons $y_1,\dots,y_p$ for linear expression of $[x_1,\dots, x_p]$ when we back-propagate the feature map in our design. Therefore, $[y_{p+1},\dots, y_q]$ can be removed. Let $W_p$ denote the square matrix $\begin{bmatrix}
 w_{11} & w_{12} & \cdots & w_{1p} \\
\vdots & \vdots & \ddots & \vdots \\
w_{p1} & w_{p2} & \cdots & w_{pp}   
\end{bmatrix}
$, we have  
\begin{equation}
\label{InverseConv}
\begin{bmatrix}
x_{1}\\
\vdots\\
x_{p}
\end{bmatrix}=W_p^{-1}
\begin{bmatrix}
y_1-b_1\\
\vdots\\
y_p-b_p
\end{bmatrix},
\end{equation}
where $W_p^{-1}$ is the inverse matrix of $W_p$. Hence, our back-propagated feature map, which is a 3D tensor, can be obtained by reshaping $[x_1,\dots,x_p]$ calculated by  Eq.~\ref{InverseConv}. 

Condition 2: When $q < p$, we consider multiple layers of $g_l$, $g_{l+1}$, ... $g_{l+M-1}$ by increasing $l$ until the number of neurons on the output of $g_{l+M-1}$ (i.e. $q_{l+M-1}$) becomes greater than or equal to $p$, for the purpose of calculating Jacobian Matrix. Next, we only select $p$ neurons $y_1^{l+M-1},\cdots,y_p^{l+M-1}$ from $q_{l+M-1}$ neurons and discard $y_{p+1}^{l+M-1},\cdots,y_q^{l+M-1}$ in the same way as proved in Condition 1. Then we have:
\begin{equation}
\label{InverseConv3}
\begin{bmatrix}
y_1^{l+M-1}\\
\vdots\\
\\
y_p^{l+M-1}
\end{bmatrix}=\begin{bmatrix}
\frac{\partial(y_{1}^{l+M-1}, \dots, y_{p}^{l+M-1})}{\partial(x_{1}^{l}, \dots, x_{p}^{l})}
\end{bmatrix}\begin{bmatrix}
x_1\\
\vdots\\
\\
x_p
\end{bmatrix}+\begin{bmatrix}
\theta_1\\
\vdots\\
\\
\theta_p
\end{bmatrix}.
\end{equation} To determine $[\theta_1,\cdots,\theta_p]$,  we can set $X=[x_1,\cdots,x_p]$ to $[0,\cdots,0]$, for obtaining $\begin{bmatrix}
\theta_1\\
\vdots\\
\theta_p
\end{bmatrix}=\begin{bmatrix}
 {y_{1}^{l+M-1}}_{|X=0}\\
 \vdots\\
 {y_{p}^{l+M-1}}_{|X=0}
\end{bmatrix}$ according to Eq.~\ref{InverseConv3}. Thus, the back-propagated feature map for convolutional layer $g_l$ can be computed by
\begin{equation}
\label{InverseConv2}
\begin{bmatrix}
x_{1}\\
\vdots\\
x_{p}
\end{bmatrix}=\begin{bmatrix}
\frac{\partial(y_{1}^{l+M-1},\dots,y_{p}^{l+M-1})}{\partial(x_{1}^{l},\dots,x_{p}^{l})}
\end{bmatrix}^{-1} \begin{bmatrix}
y_1^{l+M-1}-{y_{1}^{l+M-1}}_{|X=0}\\
\vdots\\
y_p^{l+M-1}-{y_{p}^{l+M-1}}_{|X=0}\\
\end{bmatrix}.
\end{equation}
\begin{figure*}[!t]
\centering
\includegraphics[width=1.0\textwidth]{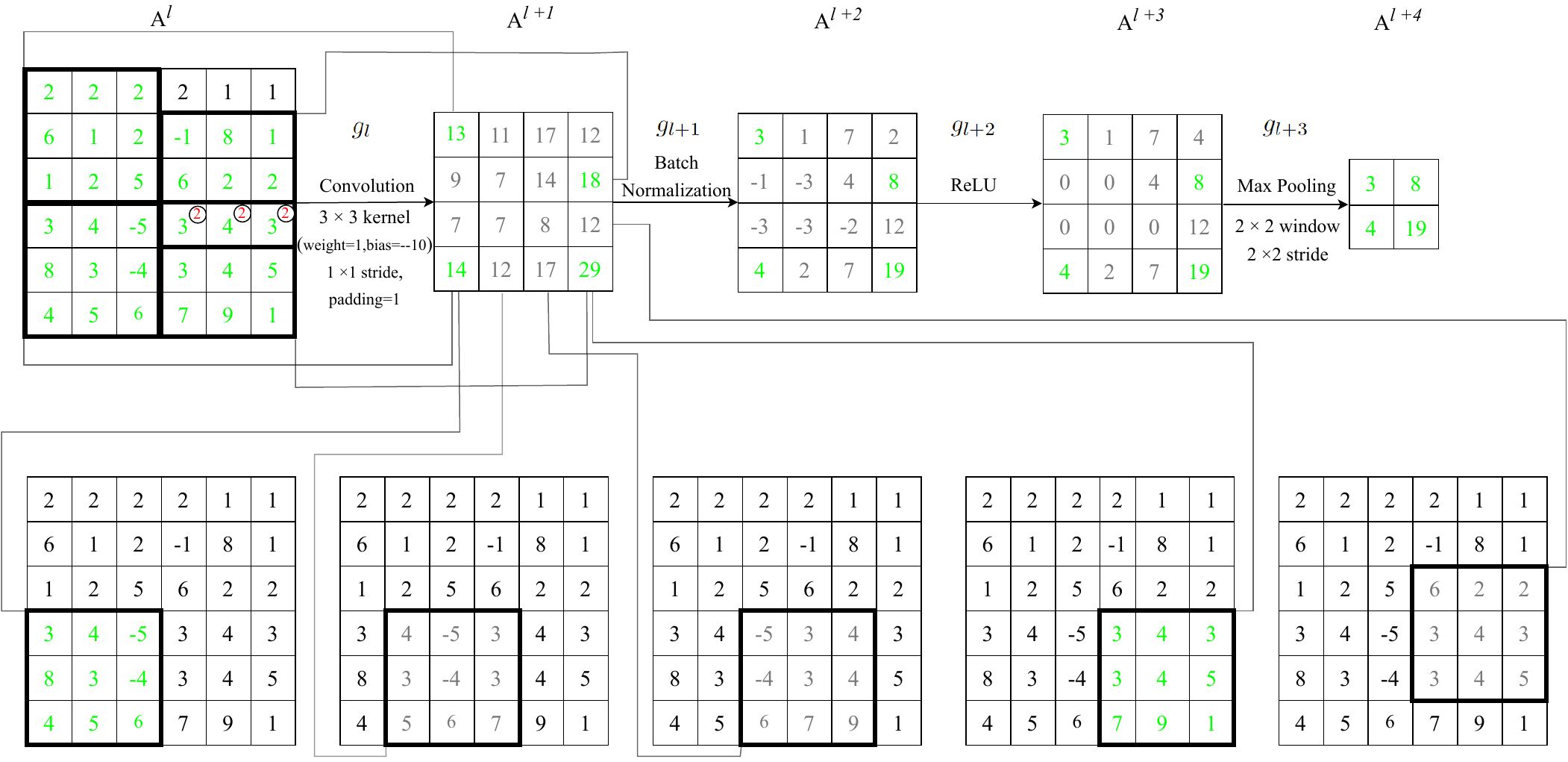}
\caption{An example illustrating the decision-making neuron-times (in green color) and abandoned neuron-times (in grey color) in four different layers of a CNN. In our proposed NA-BP, the green neuron-times are used for generating the proposed BPFMs and the grey neurons are excluded from calculating BPFMs. The number in circle $\circledS$ indicates $s$ neuron-times, that is this neuron is used $s$ times in the convolution operations of the layer.}
\label{figBPFMCNN}
\end{figure*}
\subsubsection{Back-Propagated Feature Map for Batch Normalization Layer}
When $g_l$ is a batch normalization layer, Eq.~\ref{l-layer} and Eq.~\ref{flatten} can be expressed as element-wise operation  $y_i=\beta_{i} \left(\frac{x_i-\text{Mean}}{\sqrt{\text{Var}}}\right)+\gamma_{i}$. $\beta_{i}$ and $\gamma_{i}$ are the  parameters provided by the well-trained batch normalization layer, and $\text{Mean}$ and $\text{Var}$ are the running mean and running variance respectively. Its back-propagated feature map can be obtained by reshaping $[x_1,\dots,x_p]$, which is element-wisely calculated by
\begin{equation}
\begin{aligned}
\label{InverseBN}
\text{BPFM}_l&=\text{Reshape}([x_1,\cdots,x_p]),\\
x_i=&\frac{\sqrt{\text{Var}}}{\beta_{i}}\left( y_i -\gamma_{i}\right)+\text{Mean},~i={1,\dots,p}  .
\end{aligned}    
\end{equation}
\subsubsection{Back-Propagated Feature Map for Activation Function Layer}
When $g_l$ is ReLU~\cite{refReLU}, Eq.~\ref{l-layer} and Eq.~\ref{flatten} can be expressed as the element-wise operation $y_i=
\begin{cases}
x_i, &x_i \ge 0\\
0,&x_i<0
\end{cases}
$.
The negative value neurons should not be visualized because they are not used for decision-making, which are abandoned in our back-propagated feature map calculation by reshaping $[x_1,\dots,x_p]$ that are computed by 
\begin{equation}
\begin{aligned}
\label{inverseReLU}
\text{BPFM}_l&=\text{Reshape}([x_1,\cdots,x_p]),\\
x_i=&y_i,~i={1,\dots,p}.
\end{aligned}
\end{equation}

When $g_l$ is LeakyReLU~\cite{refLeakyReLU}, Eq.~\ref{l-layer} and Eq.~\ref{flatten} are expressed as the element-wise operation $y_i=
\begin{cases}
    x_i, &x_i \ge 0\\
    \alpha \cdot x_i, &x_i<0
\end{cases}
$, where $\alpha$ is negative slope provided by LeakyReLU. Accordingly, the back-propagated feature map is obtained by reshaping $[x_1,\dots,x_p]$ that are computed by
\begin{equation}
\begin{aligned}
\label{InverseLeakyReLU}
\text{BPFM}_l&=\text{Reshape}([x_1,\cdots,x_p]),\\
x_i=&\begin{cases}
y_i,              &y_i \ge 0\\
\frac{1}{\alpha}y_i,&y_i<0
\end{cases}~. 
\end{aligned}
\end{equation}
\subsubsection{Back-Propagated Feature Map from Max Pooling Layer} When $g_l$ is a max pooling layer, the flatten operation Eq.~\ref{flatten} cannot be executed. Hence, our Eq.~\ref{l-layer} should be expressed as $(A^{l+1},\text{Index}^l)=\text{MaxPool}(A^l)$, where $\text{Index}^l$ stores the positions of selected neurons from $A^l$ by max pooling operations, while the rest of the neurons are abandoned because they are not involved in the decision-making of the model. Therefore, the back-propagated feature map consists of $x_{i}$ that is obtained by
\begin{equation}
\begin{aligned}
\label{InverseMaxpool}
\text{BPFM}_l&=\text{Reshape}([x_1,\cdots,x_p]),\\
x_{i}=&\begin{cases}
y_{i},& i \in \text{Index}^l\\
0,      &i \notin \text{Index}^l
\end{cases}~.
\end{aligned}
\end{equation}

\subsubsection{Neuron Abandoning Back-Propagation of Multiple Layers} A CNN model is constructed by stacking up several blocks with each block containing a sequence of layers, in various order and different combination, of the above listed convolutional layer, batch normalization layer, activation function layer and max pooling layer. Fig.~\ref{figBPFMCNN} gives an example showing how the Neuron Abandoning Back-Propagation (NA-BP) selects the decision-making neurons (in green) for calculating the attention maps of different internal layers of a CNN and excludes the neurons (in grey) that are not involved in making the decision. 

In the example, for the $6\times6$ region in the feature map $A^l$ fed into the convolutional layer $g_l$ which performs $16$ convolution operations, $4$ neuron blocks (each has $3\times3=9$ neurons) totalling 36 neuron-times ($33$ neurons in green with $3$ of them are used twice, which are marked by $\circled{2}$) are involved in the decision-making while the other $12$ neuron blocks totalling $12\times9=108$ neuron-times are used in the computation operations of this layer but not involved in decision-making. Note that some neurons are used multiple times in the $16$ convolution operations as the $3\times3$ kernel moves across the feature map $A^l$ as shown in the bottom row of Fig.~\ref{figBPFMCNN}. These $36$ green neuron-times are used for generating the proposed $\text{BPFM}^l$ (which will be used for generating attention map $L_l$ using Eq.~\ref{faf}) for the current convolutional layer $g_l$, while those $108$ grey neuron-times obtained during the convolution calculation are abandoned. For layer $g_{l+1}$, four neuron-times (in green) in $A^{l+1}$ contributing to the decision-making are used for generating the proposed $\text{BPFM}^{l+1}$, while $12$ grey neuron-times are discarded.

\subsection{Importance Coefficients}
When we have the back-propagated feature maps via Eqs.~\ref{BPFM},\ref{BPFM2},\ref{InverseBN},\ref{inverseReLU},\ref{InverseLeakyReLU},\ref{InverseMaxpool}, given the classification score $y^c$, the importance coefficients $\Lambda^{N}$ for $\text{BPFM}^{N}$ of the last layer $g_N$ of the CNN is calculated by
\begin{equation}
\begin{aligned}
\label{lastCoefficient}
{\Lambda}^{N}=[\lambda_1^{N},\dots,\lambda_p^{N}]=\frac{\partial y^c}{\partial [x_1^{N},\dots,x_p^{N}]}
\end{aligned},
\end{equation}
where $[x_1^{N},\dots,x_p^{N}]$ denote the neurons on $\text{A}^{N+1}$ ($\text{A}^{N+1}=\text{BPFM}^{N}$). For any internal layer of the CNN model $f^c$, the importance coefficients $\Lambda^{l}$ for $\text{BPFM}^l$ of the $l$-th layer $g_l$ are obtained by flattening the following Jacobian Matrix: 
\begin{equation}
\begin{aligned}
\label{penultimateCoefficient}
\Lambda^l=[\lambda^l_1,\dots,\lambda^l_p]=\text{Flatten}\left(\sum_{i=1}^q{\frac{\partial [\lambda_1^{l+1},\dots,\lambda_q^{l+1}]}{\partial [x_1^l,\dots,x_p^l]}}\right),\\ l=1,...,{N-1}.
\end{aligned}
\end{equation}
Repeating the calculation of $\Lambda^l$ using Eq.~\ref{penultimateCoefficient} by iteratively changing $l$ from $l=(N-1), (N-2)$ to $1$, we can obtain the importance coefficients for all the internal layers in a back-propagated manner. 

\subsection{Visualizing CNNs without Classification Score $y^c$} 
CNN models for image classification can be broadly grouped into two categories, one directly outputs classification score from FC layer, the other uses similarity metric measurement (such as those CNNs for contrastive learning image classification, few-shot image classification, image retrieval). For explaining CNN models using similarity comparison-based classification, we propose channel contribution weight in this work, as shown Fig.~\ref{fig1}, where $V_Q=[v_1^Q,v_2^Q,\cdots,v_D^Q]$ and $V_S=[v_1^{S},v_2^{S},\cdots,v_D^{S}]$ denote the feature vector of the test image and the feature vector from the feature vectors data base, respectively. 

The proposed channel contribution weight can be applied to any similarity metric. Different similarity metrics require different ways to calculate channel contribution weight. In this work, we provide formulations of computing contribution weight by taking the popular cosine similarity, that is widely used in similarity comparison-based CNN models, as an example. We compute the proposed  channel contribution weight $\Omega$ as
\begin{equation}
\begin{aligned}
\label{eqContributionWeight}
\Omega=[\omega_1,\cdots,\omega_D]=&{\frac{[{v_1^Q} {v_1^{S}},\cdots,{v_D^Q}{v_D^S}]}{{\vert{\sum_{d=1}^{D}{v_d^Q} {v_d^S}}}\vert}},
\end{aligned}
\end{equation}
where $\vert \cdot \vert$ denotes the absolute value.

\textit{Proof}: Let $\text{cos}(V_Q,V_S)$ denote the cosine similarity between two vectors, which is calculated by the equation of
\begin{equation}
\begin{aligned}
\label{CosineSimilarity}
\text{cos}(V_Q,V_S)=\frac{\sum_{d=1}^{D}v_{d}^Qv_{d}^S}{{\vert\vert V_{Q}\vert\vert}\cdot{\vert\vert V_{S}\vert\vert}}.
\end{aligned}
\end{equation}
When $\text{cos}(V_Q,V_S) \neq 0$, we will have the following,
\begin{equation}
\begin{aligned}
\label{cosine2}
1=\frac{1}{{\text{cos}(V_Q,V_S)}{\vert\vert V_Q\vert\vert}{\vert\vert V_S\vert\vert}}\sum_{d=1}^{D}v_{d}^Qv_{d}^S.
\end{aligned}
\end{equation}
According to Eq.~\ref{CosineSimilarity}, cosine similarity score $\text{cos}(V_Q,V_S)$ is the sum of $D$ items $\frac{v_d^Qv_dS}{{\vert\vert V_Q\vert\vert}{\vert\vert V_S\vert\vert}}, ~d=1,\cdots,D$, where $D$ is the number of channels. The $d$-th item $\frac{v_d^Qv_dS}{{\vert\vert V_Q\vert\vert}{\vert\vert V_S\vert\vert}}$ represents the contribution to the similarity score from the $d$-th channel. To measure channel-wise contribution in percentage, we can determine and calculate the $d$-th channel contribution weight as 
\begin{equation}
\begin{aligned}
\label{omiga}
\omega_d=&\frac{v_d^Qv_d^S}{{\vert\text{cos}(V_Q,V_S)\vert}{\vert\vert V_Q\vert\vert}{\vert\vert V_S\vert\vert}}\\
=&\frac{v_d^Qv_d^S}{\vert{\sum_{d=1}^{D}{v_d^Q} {v_d^S}}\vert}.
\end{aligned}  
\end{equation}

The proposed channel contribution weight $\omega_d$ can measure the importance of the $d$-th channel to the similarity metrics-based classification decision. Therefore, the importance coefficients for $\text{BPFM}^N$ are calculated by the following Jacobian Matrix,
\begin{equation}
\begin{aligned}
\label{eqImportCoefficients}
\Lambda^N=[\lambda^N_1,\dots,\lambda^N_p]=\text{Flatten}\left(\sum_{i=1}^D{\frac{\partial [\omega_1,\dots,\omega_D]}{\partial [x_1^N,\dots,x_p^N]}}\right).
\end{aligned}
\end{equation}

\begin{figure*}[!t]
\centering
\includegraphics[width=1.0\textwidth]{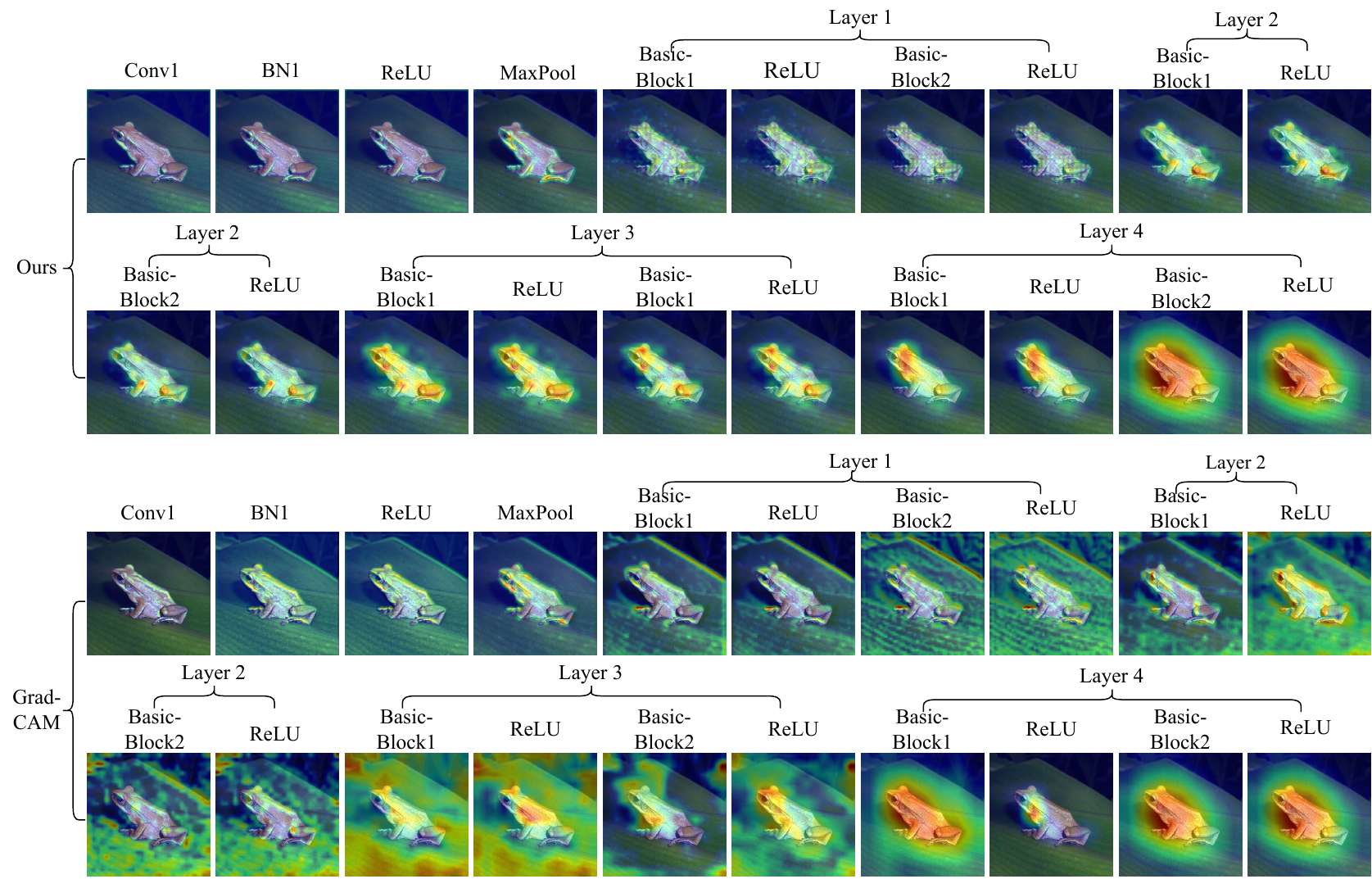}
\caption{The attention maps for all the layers of ResNet18 visualised by the proposed NAFlow in comparison with that generated by the Grad-CAM (using the public code library$^2$ to generate the attention maps for intermediate layers). Note that the attention maps for the final output layer of ResNet18 generated by both the proposed NAFlow and the Grad-CAM are same, while the attention maps for the intermediate layers are different. The reported accuracy of ResNet18 on ImageNet2012 is $72.12\%$~\cite{refResNet}.}
\label{fig8}
\end{figure*}
\section{Experiments}
\subsection{CNN Models}
Nine publicly released CNN models including 4 widely-used CNN models with FC layer as classifier for general image classification (ResNet18~\cite{refResNet}, ResNet50~\cite{refResNet}, ResNet50$+$CBAM~\cite{refCBAM}, and ResNet50$+$BAM~\cite{BAM}) and 5 similarity metric based CNN models for contrastive learning classification (ResNet18-NPID~\cite{refNPID}, ResNet50-NPID~\cite{refNPID}, few-shot image classification (Conv64FF~\cite{refConv64F}, Conv4Net~\cite{refPNet}), and image retrieval (ResNet50-MS~\cite{refMSLoss}) are used for evaluating the effectiveness of the proposed method. Among the 9 CNN models, two of them (Conv4Net, and Conv64F) use LeakyReLU~\cite{refLeakyReLU} as the activation function layer while ReLU~\cite{refReLU} is employed as the activation function layer in the other seven CNNs of ResNet18,  ResNet50, ResNet50$+$CBAM, ResNet50$+$BAM, ResNet18-NPID, ResNet50-NPID, and ResNet50-MS. 
\subsection{Dataset and Experiment Setting}
The experiments are conducted on the ImageNet2012 dataset ~\cite{refImageNet2012} and the fine-grained image dataset CUB200~\cite{refCUB200} as used by the nine CNN models. The ImageNet dataset contains around 1.3 million training images and 50,000 images in validation set, labelled across 1,000 semantic categories. The CUB200 includes 11,788 images from 200 classes. 

Before feeding into the CNN models, all images are scaled into $[0,1]$ and then normalized by using using mean $[0.485,0.456, 0.406]$ and standard deviation $[0.229,0.224, 0.225]$. All our experiments are conducted by using Pytorch library with Python 3.8 on a NVIDIA RTX 3090 GPU.

\begin{figure*}[!t]
\centering
\includegraphics[width=1.0\textwidth]{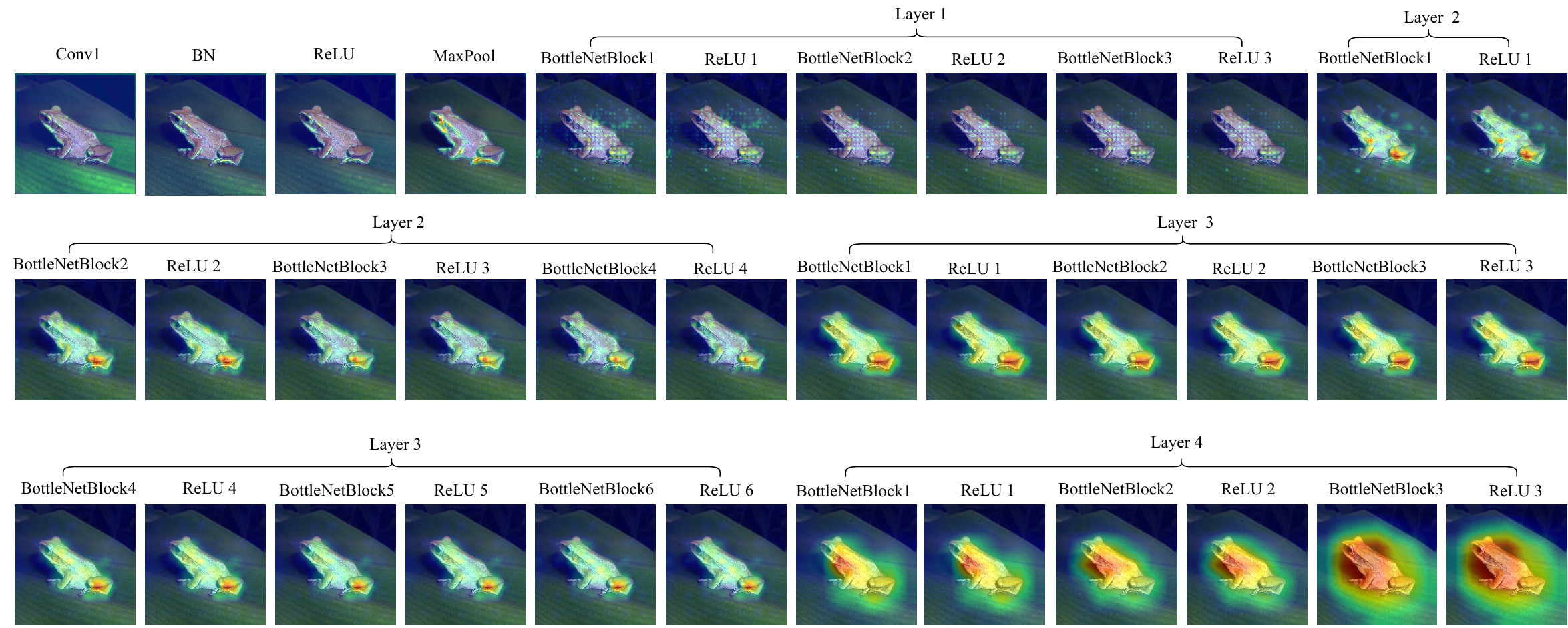}
\caption{The attention maps for all layers of ResNet50 visualized by the proposed NAFlow. The reported accuracy of ResNet50 on ImageNet2012 is $79.26\%$~\cite{refResNet}, higher than that of ResNet18 shown in Fig.~\ref{fig8}.}
\label{resnet50fcfrog}
\end{figure*}
\begin{figure*}[!t]
\centering
\includegraphics[width=1.0\textwidth]{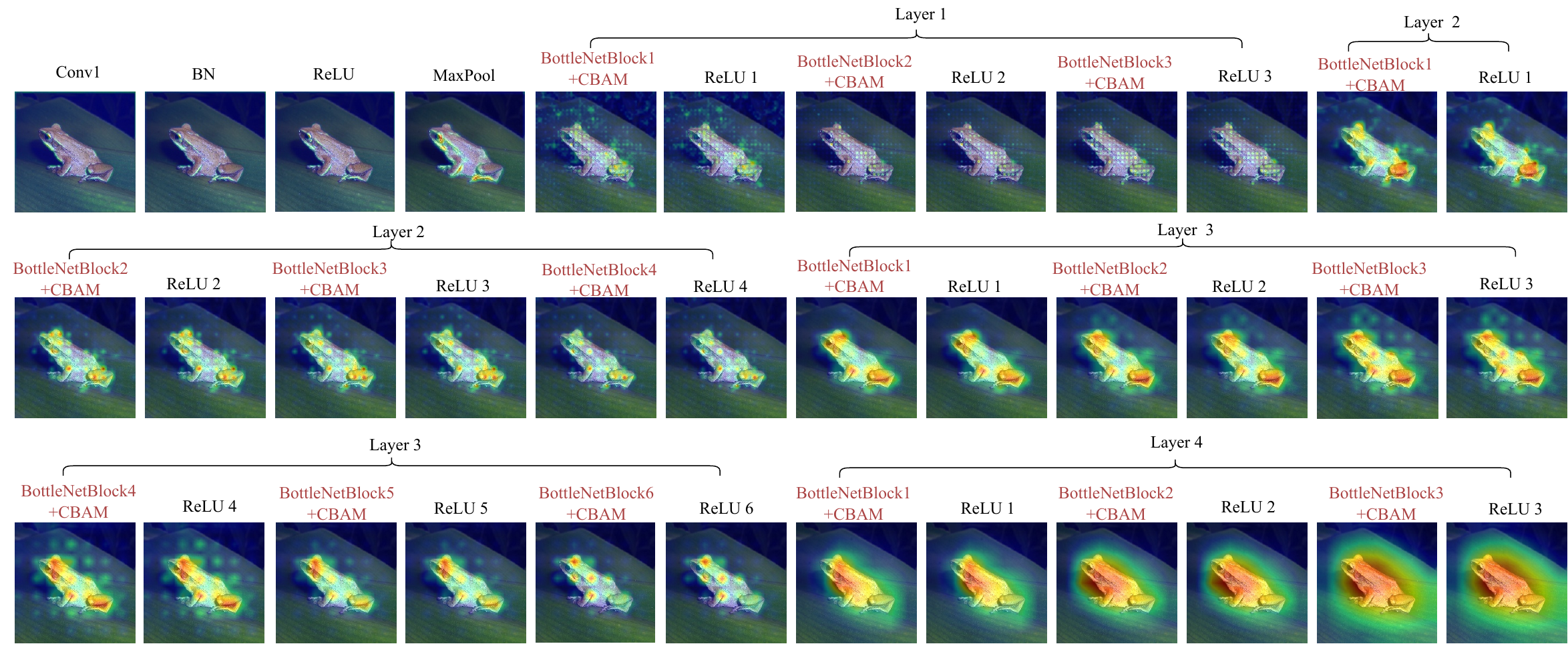}
\caption{\textcolor{black}{The attention maps for all layers of ResNet50+CBAM visualized by the proposed NAFlow. The CBAM layers are highlighted in red. The reported accuracy of ResNet50-CBAM on ImageNet2012 is $77.34\%$~\cite{refCBAM} while their reported accuracy of ResNet50 is $75.44\%$~\cite{refCBAM}.}}
\label{resnet50cbamfcfrog}
\end{figure*}
\begin{figure*}[!t]
\centering
\includegraphics[width=1.0\textwidth]{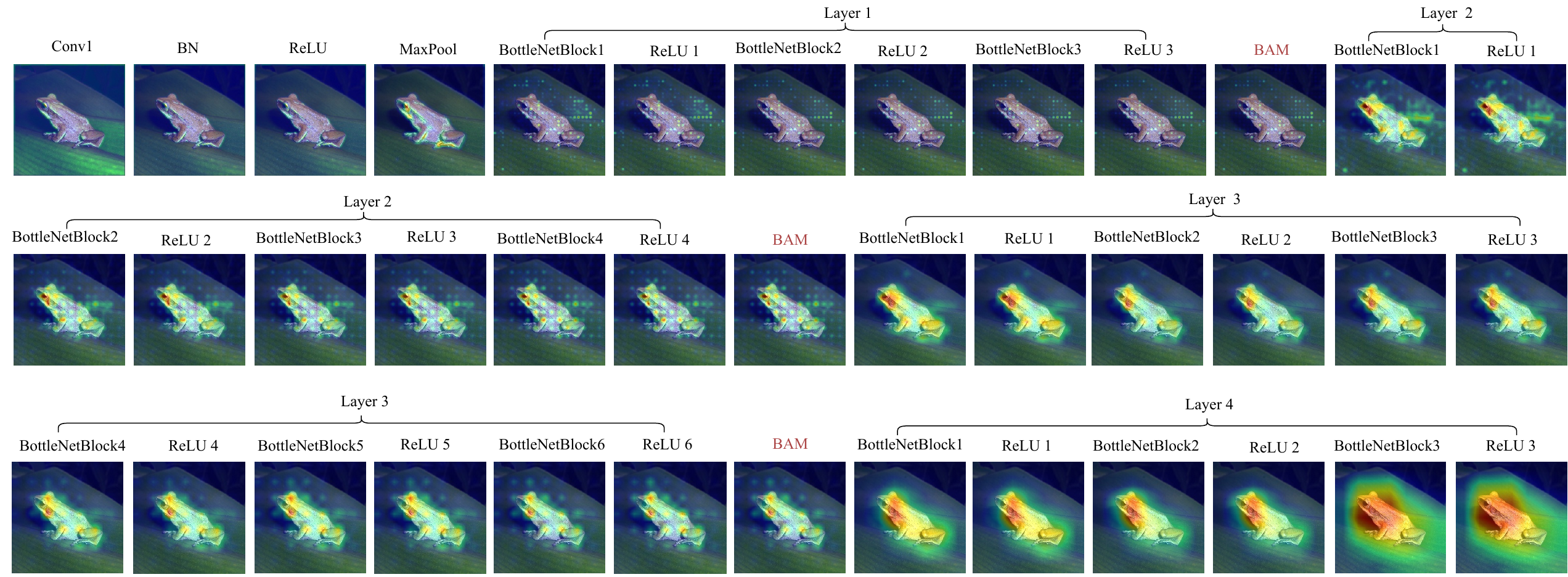}
\caption{\textcolor{black}{The attention maps for all layers of ResNet50+BAM visualized by the proposed NAFlow. The additional Layers of BAM are highlighted in red. The reported accuracy of ResNet50-BAM on ImageNet2012 is $75.98\%$~\cite{BAM} while their reported accuracy of ResNet50 is $75.44\%$~\cite{BAM}.}}
\label{resnet50bamfcfrog}
\end{figure*}


\begin{figure*}[t]
\centering
\includegraphics[width=1.0\textwidth]{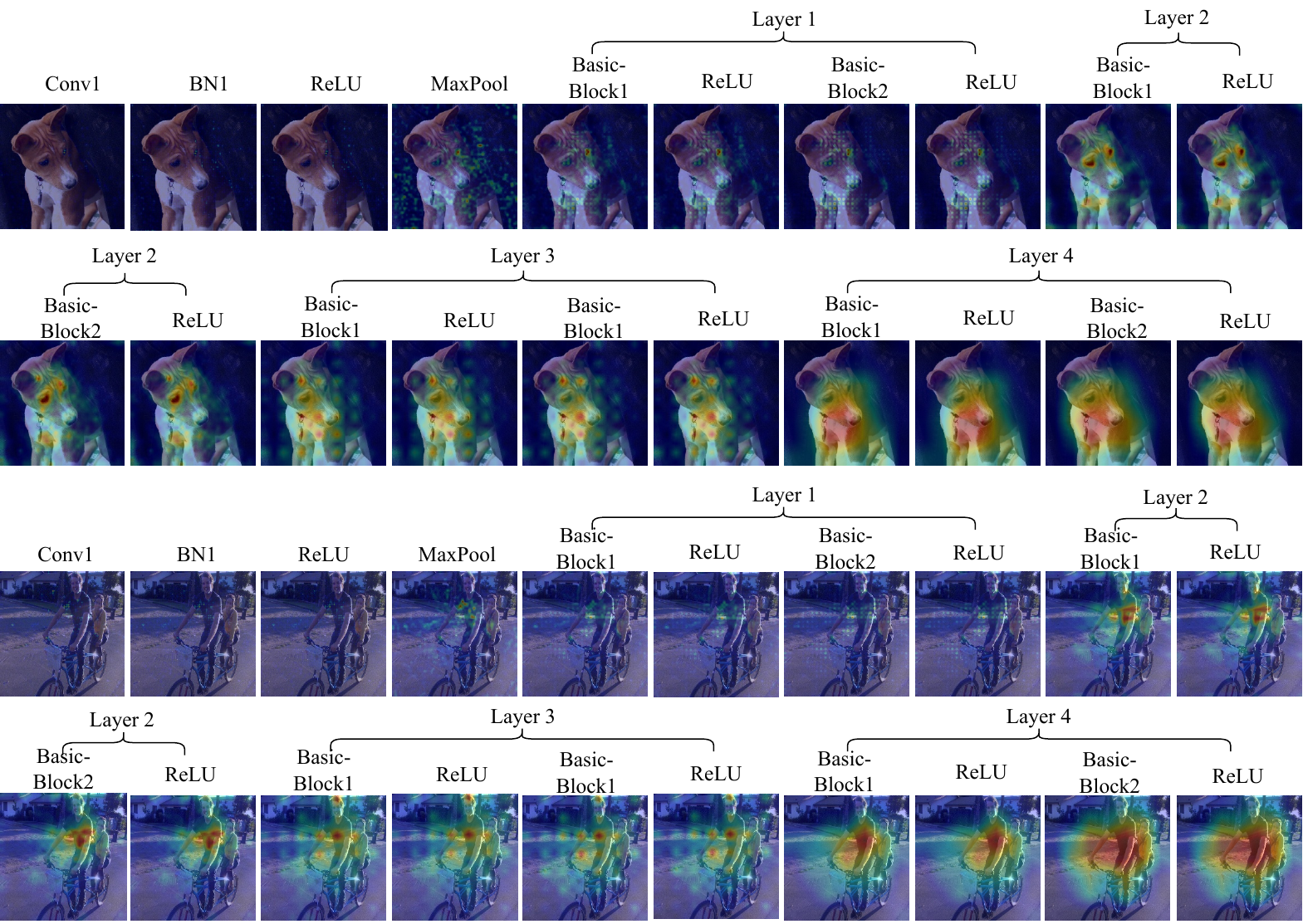}
\caption{The attention map evolution over all layers of ResNet18-NPID (a CNN model for contrastive learning image classification) visualized by the proposed NAFlow on an example image from the ImageNet2012. The reported accuracy of ResNet18-NPID on ImageNet2012 is $41\%$~\cite{refNPID}. The top rows show an example of correctly classified images by the ResNet18-NPID, while the botton image of tandem bicycle is misclassified as athlete by the ResNet18-NPID.}
\label{fig5}
\end{figure*}

\begin{figure*}[!t]
\centering
\includegraphics[width=1.0\textwidth]{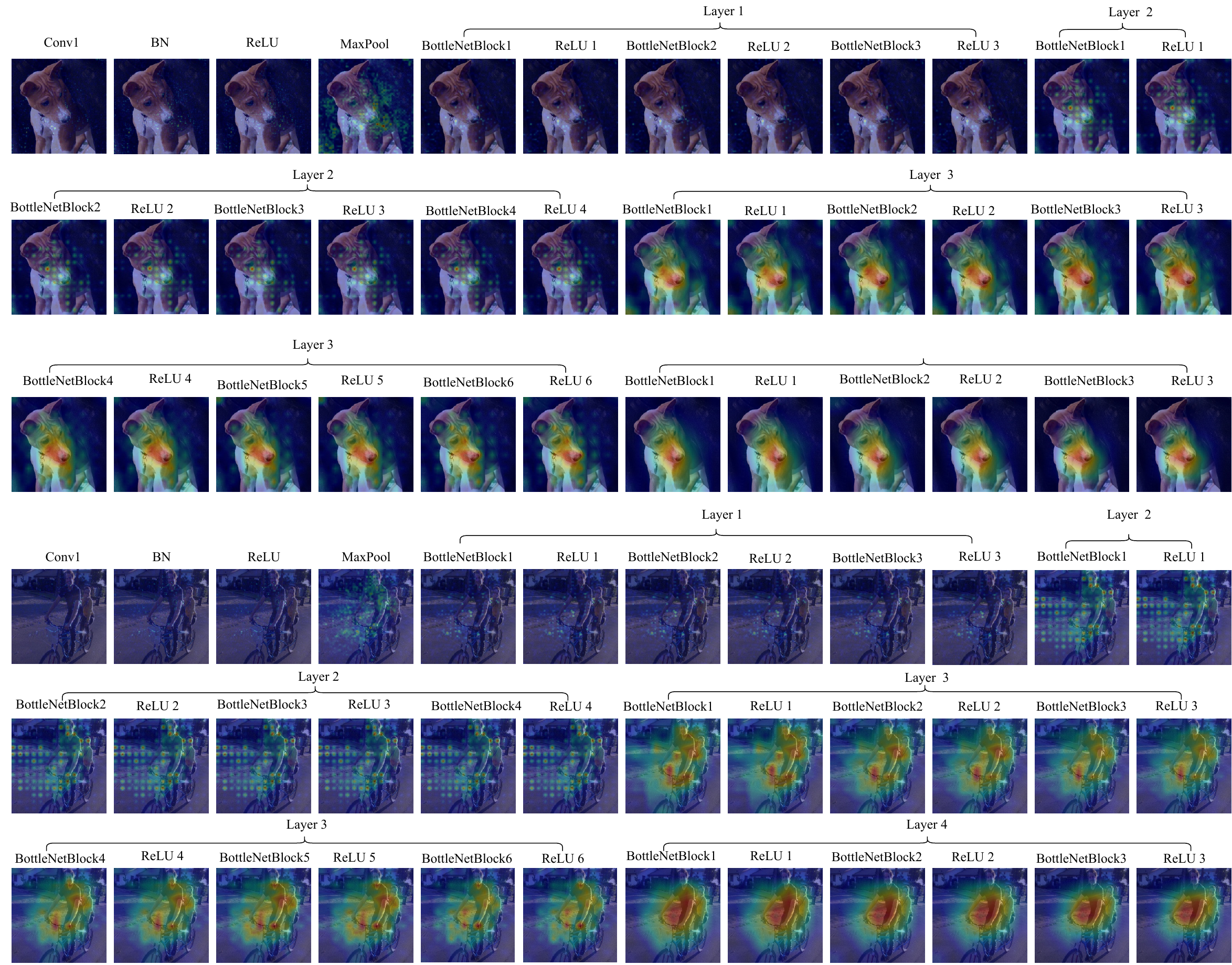}
\caption{The attention flow moving through all layers of ResNet50-NPID (another CNN model for contrastive learning image classification) visualized by the proposed NAFlow on an example image from the ImageNet2012. The reported accuracy of ResNet50-NPID on ImageNet2012 is $46\%$~\cite{refNPID}. Both dog and tandem bicycle images in Fig.~\ref{fig5} are correctly classified by the ResNet50-NPID.}
\label{fig6}
\end{figure*}



\begin{figure*}[!t]
\centering
\includegraphics[width=1.0\textwidth]{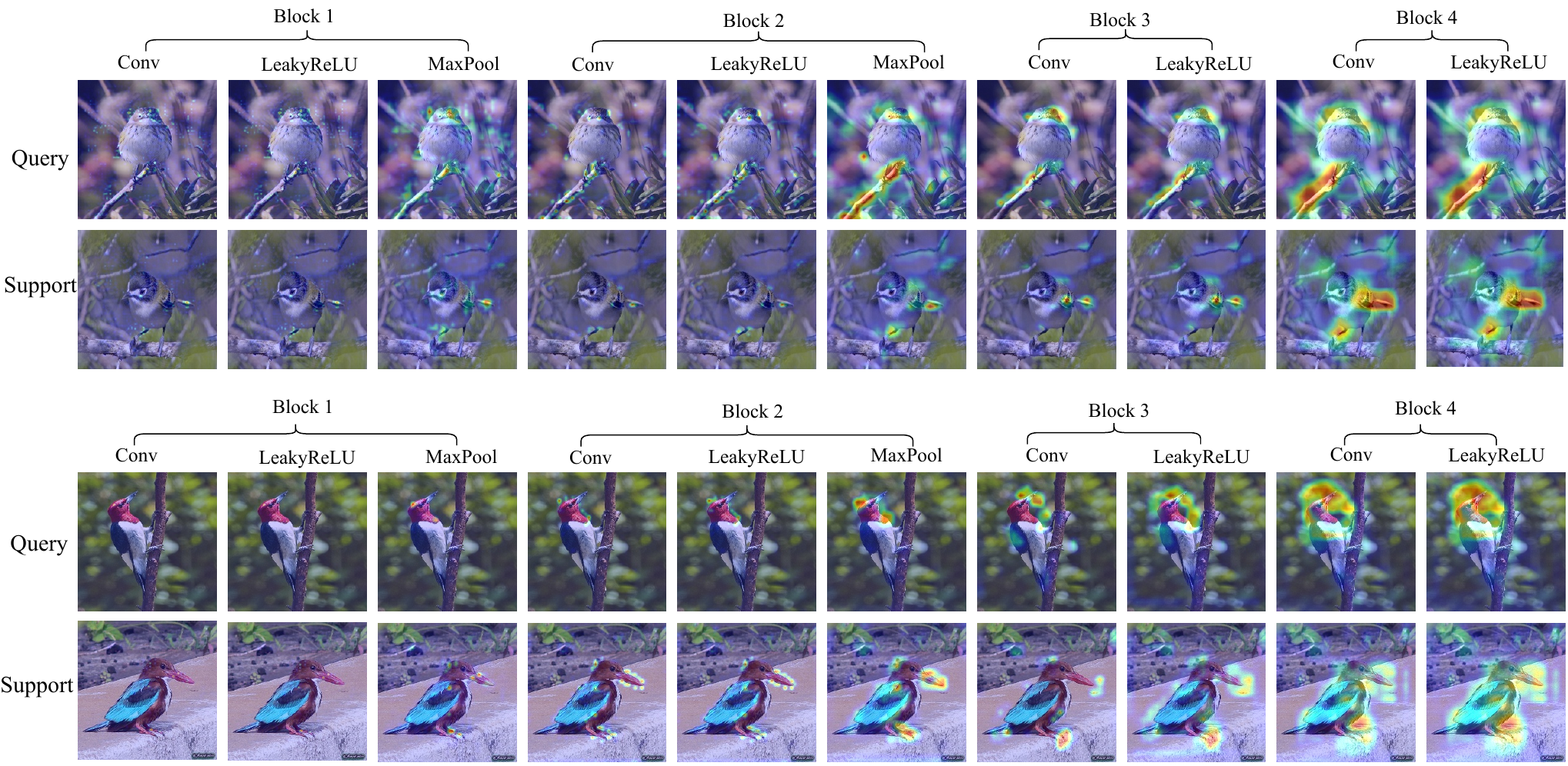}
\caption{Example attention flows moving through all layers of Conv64F visualized by the proposed NAFlow for 5-way 1-shot image classification on CUB200 dataset. The reported average accuracy over 10,000 episodes is $62.42\%$~\cite{refMetaBaseline}. These two query images of Myrtle Warbler and Red Headed Woodpecker are misclassified by Conv64F into other classes of  Black Capped Vireo and White Breasted Kingfisher respectively as shown by the support images, which are however correctly matched by Conv4Net as shown in Fig.~\ref{fig2}.}
\label{fig4}
\end{figure*}
\begin{figure*}[!t]
\centering
\includegraphics[width=1.0\textwidth]{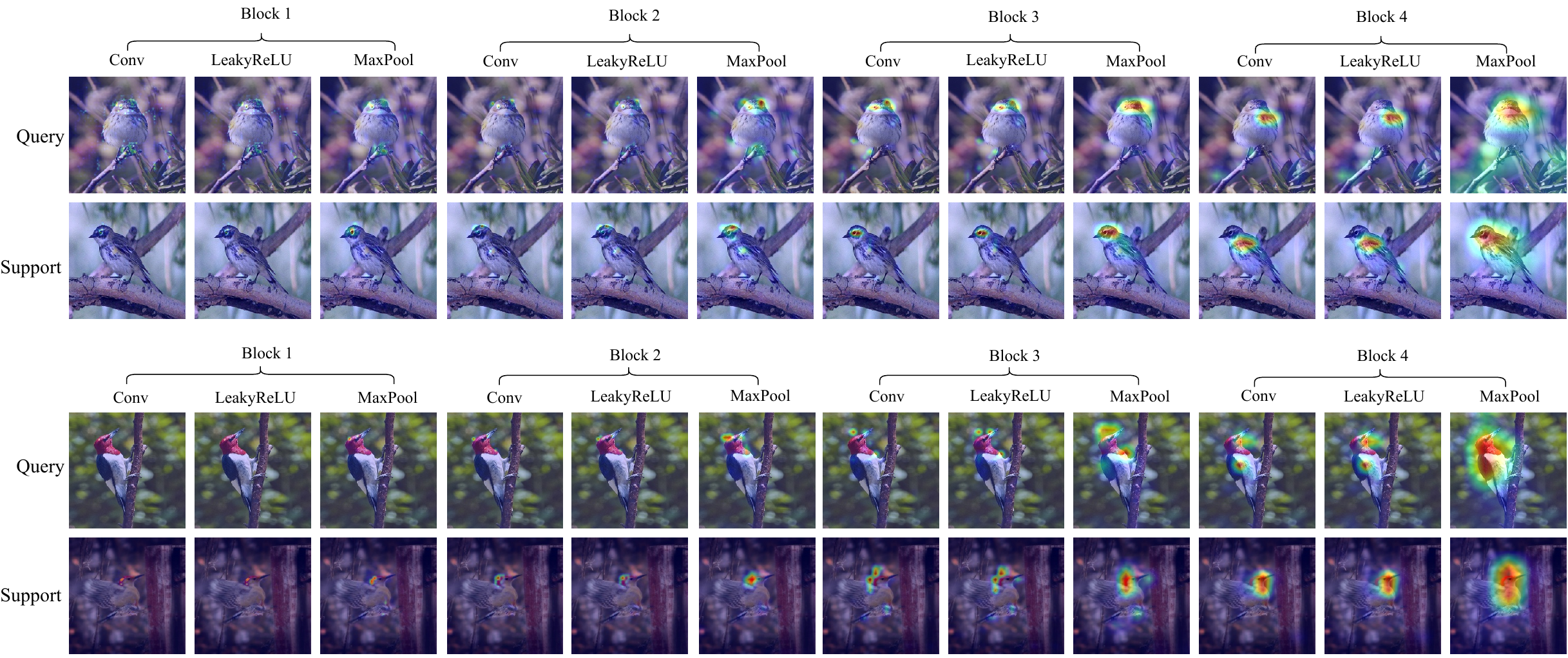}
\caption{Example attention flows moving through all layers of Conv4Net visualized by the proposed NAFlow for 5-way 1-shot image classification on CUB200 dataset. The reported average accuracy over 10,000 episodes is $65.88\%$~\cite{refMetaBaseline}.}
\label{fig2}
\end{figure*}

\begin{figure*}[!t]
\centering
\includegraphics[width=1.0\textwidth]{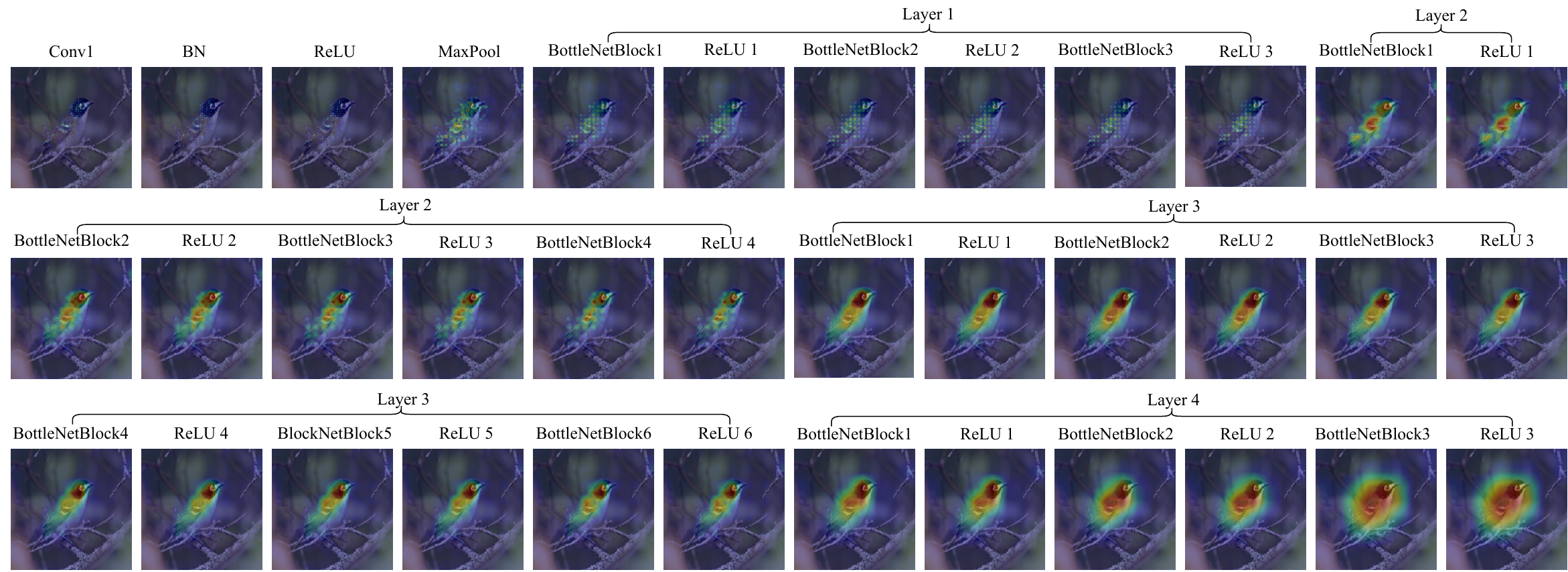}
\caption{An example attention flows moving through all layers of ResNet50-MS on
CUB200 dataset visualized by the proposed NAFlow. The model's reported Recall@$1$ on
CUB200 dataset is $65.7\%$~\cite{refMSLoss}.}
\label{fig7}
\end{figure*}
\subsection{The Neuron Abandoning Attention Flow for Visualizing Evolution of Attentions inside CNN Models}
We first examine the proposed method on the two most widely used CNN models for general image classification, ResNet18~\cite{refResNet} and ResNet50~\cite{refResNet}, which was pretrained on ImageNet2012 training set. The publicly released codes and the parameters of their pretrained models\footnote{https://github.com/pytorch/examples/tree/main/imagenet} are used, which have reported accuracies of $72.12\%$ and $79.26\%$ respectively on the validation set of ImageNet2012. The example sequences of attention maps from internal to output layers visualized by the proposed NAFlow on an image randomly selected from the testing samples correctly predicted by ResNet18 and ResNet50 are shown in Fig.~\ref{fig8} and Fig.~\ref{resnet50fcfrog} respectively.

As explained in Sections 1 and 2, all the existing
visual explanation methods can only be used for generating
the attention map of the final output layer but not for the
internal layers of CNN models. Using them to generate
attention maps for the intermediate layers is conceptually incorrect, containing large amount of feature information from
irrelevant neurons not contributing to the decision-making of
the CNN. In Fig.~\ref{fig8}, we also display the attentions maps for internal layers generated by Grad-CAM using the widely-used library\footnote{https://github.com/jacobgil/pytorch-grad-cam}. Fig.~\ref{fig8} and Fig.~\ref{resnet50fcfrog} show that both ResNet18 and Resnet50 have their attentions located on the edges of objects and also scattered on the background initially. When the layers go deeper, the attentions gradually extend to the whole body and their intensity becomes stronger. At the same time, those attentions scattered on the background not only do not grow bigger in size but their intensities also decreases. These results indicate that the two models look functioning close to perfect and their publicly released parameters are well trained. However, the attentions for the internal layers generated by Grad-CAM are scattered randomly when moving from shallow layers to deeper layers (see the bottom two rows in Fig.~\ref{fig8}) because it uses all the neurons in each internal layer, showing the errors caused by the irrelevant neurons that do not contribute to the decision-making. The attention maps for the last output layer generated by the proposed NAFlow and Grad-CAM are same as all neurons in this layer contribute to making decision and no neuron is abandoned by NAFlow.

\textcolor{black}{We also use the proposed NAFlow method as an analytical tool to analyse another two CNN models for general image classification, ResNet50+CBAM~\cite{refCBAM} and ResNet50+BAM~\cite{BAM}, particularly justifying the effectiveness of any module integrated into CNN architectures in terms of the attention capturing. Their publicly released codes and
the parameters of the pretrained models\footnote{https://github.com/Jongchan/attention-module} are used, which
have reported accuracies of $77.34\%$ and $75.98\%$ respectively on the validation set of ImageNet2012. The Convolutional Block Attention Module (CBAM)~\cite{refCBAM} and Bottleneck Attention Module (BAM)~\cite{BAM} are modules that are plugged into a CNN model for accuracy improvement. ResNet50+CBAM and ResNet50+BAM are CNNs by integrating CBAM and BAM into ResNet50 that have achieved reported accuracy improvements of $1.90\%$ (from $75.44\%$ to $77.34\%$)~\cite{refCBAM} and $0.54\%$ (from $75.44\%$ to $75.98\%$)~\cite{BAM}. To understand what happened inside the CNNs after CBAM and BAM are inserted, the proposed NAFlow is applied to ResNet50-CBAM and ResNet50-BAM to generate the attention flow, which is shown in Fig.~\ref{resnet50cbamfcfrog} and Fig.~\ref{resnet50bamfcfrog} respectively. By comparing with Fig.~\ref{resnet50fcfrog}, it is observed that the  BottleNetBlock integrated with CBAM (BottleNetBlock+CBAM) is able to spread the attention wider and later  focus on more regions on the target when the attention moves from shallow layers to deeper layers in the CNN model. From Fig.~\ref{resnet50bamfcfrog}, the inserted BAM modules in the ResNet50+BAM model breakup the attention in the 8 layers of $``\text{layer}~2"$ and spread the attention over more parts (e.g., head) when it moves to deeper layers. It is interesting to see that the attention maps of the final output layer of ResNet50-CBAM and ResNet50-BAM are similar to that of ResNet50, and the function of CBAM and BAM are easily revealed from internal attention flow but not easily by the ``external" attention map of the last output layer. In this case, NAFLow provides an analysis tool to look into the dynamics inside CNN models similar to that medical CT-scans provide powerful internal diagnostic evidence although the symptoms are not differentiable from outside appearance of patients.}

Next, we apply the proposed NAFlow to visualize two contrastive learning classification CNN models, ResNet18-NPID~\cite{refNPID} and ResNet50-NPID~\cite{refNPID}. The publicly released codes and the parameters of their pretrained models\footnote{http://github.com/zhirongw/lemniscate.pytorch} are used, whose reported accuracies are $41.0\%$ and $46.8\%$ respectively. The example sequences
of attention maps from internal to output layers visualized by
the proposed NAFlow on  images randomly selected from the
testing samples correctly and incorrectly predicted by the two contrastive learning image classification models of ResNet18-NPID and ResNet50-NPID are shown in Fig.~\ref{fig5} and Fig.~\ref{fig6}. The ResNet18-NPID correctly classified the dog image and its attention maps in Fig.~\ref{fig5} look well located and gradually stabilized on the dog. However for the more complex tandem bicycle image, ResNet18-NPID's attention maps in NAFlow look not yet stabilized from shallow internal layers to the output layer. This indicates the model architecture and parameters of ResNet18-NPID are capable to classify the dog image but not enough to handle more complex contents in the tandem bicycle image. In comparison with Fig.~\ref{fig6}, the attention maps in NAFLow show ResNet50-NPID's attention has stabilized over layers on the objects of both images, showing the architecture and the pretrained parameters of ResNet50-NPID reach the level of handling the complexity in the tandem bicycle image that ResNet18-NPID cannot. 

We further apply the proposed NAFlow to visually explain another type of CNN models, few-shot image classification CNNs. The publicly released codes\footnote{https://github.com/yinboc/few-shot-meta-baseline} of two popular few-shot CNNs (Conv64F~\cite{refConv64F}, Conv4Net~\cite{refPNet}) are used for our experiment. The default settings of the codes are used, which include 5-way 1-shot episode scheme using MetaBaseline training method~\cite{refMetaBaseline}, using cosine similarity as the similarity metric, and the maximum of 5 similarity scores between 1 query and 5 support images are considered as the prediction. The examples of attention maps from internal to output layers
visualized by the proposed NAFlow on images randomly selected from the testing samples incorrectly predicted by Conv64F but correctly predicted by Conv4Net are shown in Fig.~\ref{fig4} and  Fig.~\ref{fig2}, respectively. In Fig.~\ref{fig4}, it is observed that Conv64F incorrectly matched the query images of Myrtle Warbler and Red Headed Woodpecker to the support images of Black Capped Vireo and White Breasted Kingfisher respectively. Their attentions in all the layers are always at different regions between the matched query and support images. On contrast, the same two query images are correctly predicted by Conv4Net as shown by Fig.~\ref{fig2}. The attention of Conv4Net initially starts to focus on the head of the bird (Myrtle Warbler) at the layer 3 till the layer 9, then moves on to the chest of the bird at the layers 10 and 11. It finally enlarges to include both head and chest. Both query and support images show the same pattern of attention changes, which does not exist in Fig.~\ref{fig4}. This explains from the attention flow perspective why Conv4Net makes a correct prediction on these query images while Conv64F does not. 

Finally, we applied the proposed NAFlow to the ResNet50-MS model that is used for image retrieval task. The publicly released code and pretrained parameters\footnote{https://github.com/msight-tech/research-ms-loss} of ResNet50-MS are used. The pretrained model achieves the Recall@$1$ of $65.7\%$~\cite{refMSLoss}. Fig.~\ref{fig7} shows the NAFlow of an example image that is correctly retrieved by the CNN model. Similar to all the previous eight CNN models for general image classification, contrastive learning classification, and few-shot image classification tasks, the proposed NAFLow demonstrates consistent performance and effectiveness on the CNN for another type of task, i.e., image retrieval.  



\section{Conclusion}
Current methods for visually explaining CNN models are only able to interpret the output layer attention of a CNN when making its decision. This study attempts to look deep inside a CNN model to decipher how the model progressively forms, from layer to layer, its decision. A novel neuron abandoning attention flow method is proposed that can trace and find those neurons that are involved in the decision-making using a neuron abandoning back propagation strategy. Via generating the back-propagated feature maps and tensors of importance coefficients by using inverse function of intermediate layers of the CNN model, we accurately locate decision-making neurons (or neuron-times) for every internal layer to construct attention flow through all layers of the model. This work also fills a missing gap of not able to visually explaining CNNs using similarity metric based classification by proposing a new method of Jacobian matrix computation of channel contribution weights. Extensive experiments on nine CNN models for different tasks of general image classification, contrastive learning image classification, few-shot image classification, and image retrieval demonstrate the consistent effectiveness of NAFlow and its potential as a more insightful tool in explaining CNNs, analysing effectiveness of internal components of a model, and guide the design of new models.








\bibliography{NAFlow}
\bibliographystyle{IEEEtran}


\vspace{11pt}

\begin{IEEEbiography}[{\includegraphics[width=1in,height=1.25in,clip,keepaspectratio]{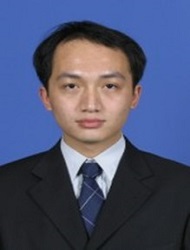}}]
{Yi Liao} received the B.S. degree in clinical medicine from Fudan University, Shanghai, China, in 2007, and M.S. degree of computer science from Queensland University of Technology, Brisbane, Australia, in 2019. He is currently pursuing the Ph.D. degree of Artificial Intelligence at School of Engineering and Built Environment, Griffith University, Brisbane, Australia. 

His research interests include deep learning, image classification and object recognition.
\end{IEEEbiography}
\begin{IEEEbiography}
[{\includegraphics[width=1in,height=1.25in,clip,keepaspectratio]{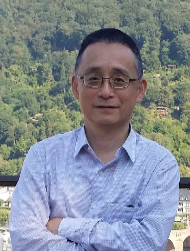}}]{Yongsheng Gao} received the B.Sc. and M.Sc. degrees in electronic engineering from Zhejiang University, Hangzhou, China, in 1985 and 1988, respectively, and the Ph.D. degree in computer engineering from Nanyang Technological University, Singapore. He is currently a Professor with the School of Engineering and Built Environment, Griffith University, and the Director of ARC Research Hub for Driving Farming Productivity and Disease Prevention, Australia. He had been the Leader of Biosecurity Group, Queensland Research Laboratory, National ICT Australia (ARC Centre of Excellence), a consultant of Panasonic Singapore Laboratories, and an Assistant Professor in the School of Computer Engineering, Nanyang Technological University, Singapore. 

His research interests include smart farming, machine vision for agriculture, biosecurity, face recognition, biometrics, image retrieval, computer vision, pattern recognition, environmental informatics, and medical imaging. 
\end{IEEEbiography}
\begin{IEEEbiography}
[{\includegraphics[width=1in,height=1.25in,clip,keepaspectratio]{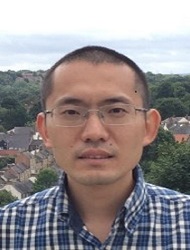}}]{Weichuan Zhang}
received the MS degree in signal and information processing from the Southwest Jiaotong University in China and the PhD degree in signal and information processing in National Lab of Radar Signal Processing, Xidian University, China. He is currently a Professor with the School of Electronic Information and Artificial Intelligence, Shaanxi University of Science and Technology, Xi 'an, Shaanxi Province, China, and a Principal Research Fellow at Griffith University, QLD, Australia. 

His research interests include computer vision, image analysis, and pattern recognition. He is a member of the IEEE.
\end{IEEEbiography}

\vfill

\end{document}